\documentclass{article}
\usepackage{ijcai25}

\usepackage{times}
\usepackage{soul}
\usepackage{url}
\usepackage[hidelinks]{hyperref}
\usepackage[utf8]{inputenc}
\usepackage[small]{caption}
\usepackage{graphicx}
\usepackage{amsmath}
\usepackage{amsthm}
\usepackage{booktabs}
\usepackage{algorithm}
\usepackage{algorithmic}
\usepackage[switch]{lineno}

\usepackage{dblfloatfix}
\usepackage{amssymb}
\usepackage{caption}
\usepackage{subcaption}
\usepackage{graphicx}

\title{HanDrawer: Leveraging Spatial Information to Render Realistic Hands Using a Conditional Diffusion Model in Single Stage}

\author{
Qifan Fu$^{1,2}$
\and
Xu Chen$^{1,3}$\and
Muhammad Asad$^1$\and
Shanxin Yuan$^{1,2}$\and
Changjae Oh$^2$\And
Gregory Slabaugh$^{1,2}$\\
\affiliations
$^1$Digital Environment Research Institute, Queen Mary University of London, London, UK\\
$^2$School of Electronic Engineering and Computer Science, Queen Mary University of London, London, UK\\
$^3$Department of Medicine, University of Cambridge, Cambridge, UK\\
\emails
q.fu@qmul.ac.uk
}

\begin{document}
\maketitle
\begin{abstract}
Although diffusion methods excel in text-to-image generation, generating accurate hand gestures remains a major challenge, resulting in severe artifacts, such as incorrect number of fingers or unnatural gestures. 
To enable the diffusion model to learn spatial information to improve the quality of the hands generated, we propose \emph{HanDrawer}, a module to condition the hand generation process. Specifically, we apply graph convolutional layers to extract the endogenous spatial structure and physical constraints implicit in MANO hand mesh vertices. We then align and fuse these spatial features with other modalities via cross-attention. The spatially fused features are used to guide a single stage diffusion model denoising process for high quality generation of the hand region. To improve the accuracy of spatial feature fusion, we propose a Position-Preserving Zero Padding (PPZP) fusion strategy, which ensures that the features extracted by HanDrawer are fused into the region of interest in the relevant layers of the diffusion model. 
HanDrawer learns the entire image features while paying special attention to the hand region thanks to an additional hand reconstruction loss combined with the denoising loss. 
To accurately train and evaluate our approach, we perform careful cleansing and relabeling of the widely used HaGRID hand gesture dataset and obtain high quality multimodal data. Quantitative and qualitative analyses demonstrate the state-of-the-art performance of our method on the HaGRID dataset through multiple evaluation metrics. \emph{Source code and our enhanced dataset will be released publicly if the paper is accepted.}
\end{abstract}

\section{Introduction}

\begin{figure}[t]
    \centering
    \includegraphics[width=\columnwidth]{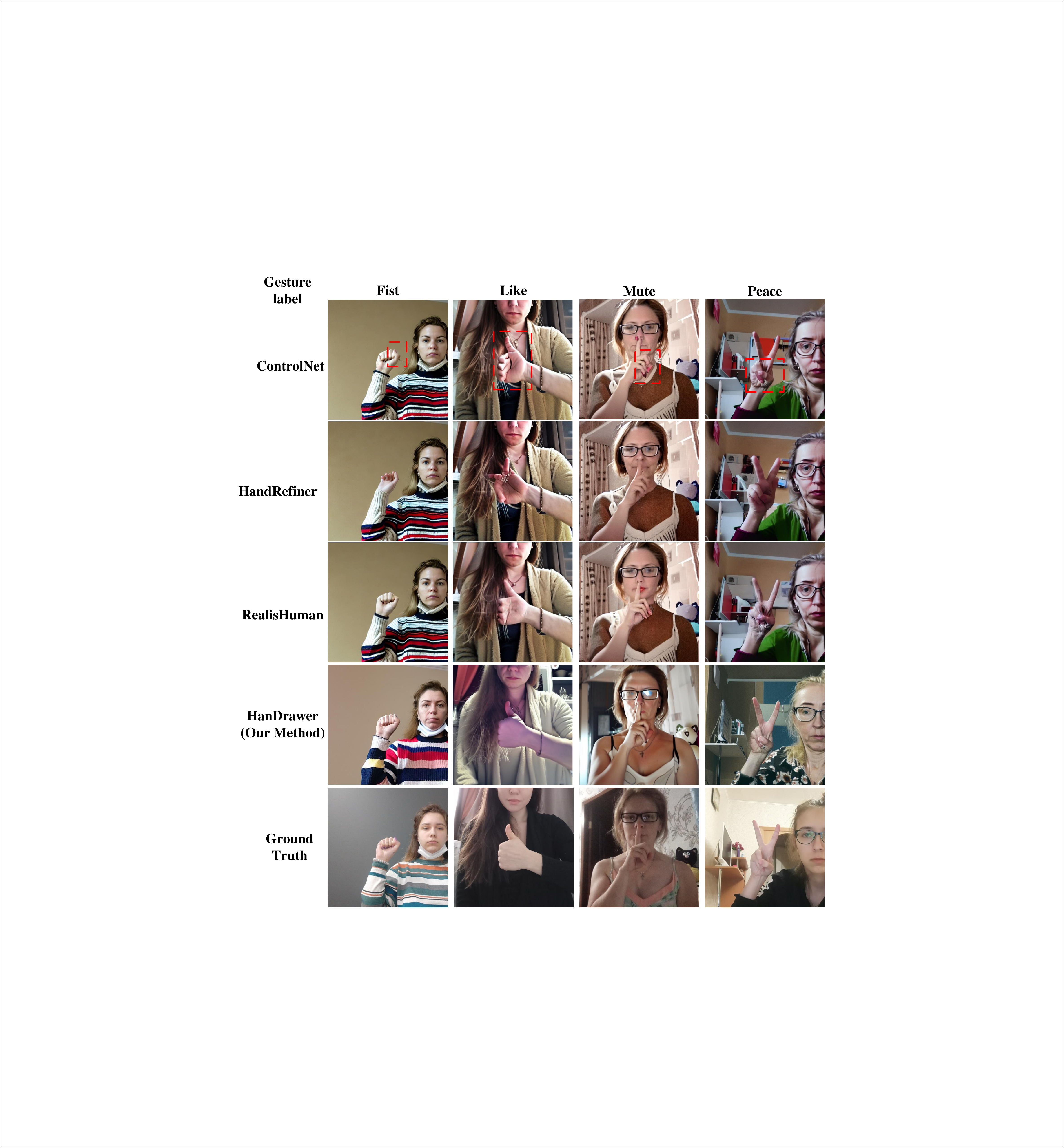}
    \caption{\textbf{Accurate and realistic hand gesture generation.} ControlNet~\protect\cite{zhang2023adding} can guide Stable Diffusion to generate accurate body poses, but the precision and realism of the generated hand gestures are quite poor (first row). Our method (fourth row) improves the precision and realism of generated gestures by learning the complex intrinsic structure and spatial information of gestures to better match the ground truth pose (bottom row) compared to other methods (second and third rows).}
    \label{fig:FIG1}
\end{figure}

Diffusion models have shown stunning performance on text-to-image generation, and since their introduction, various control methods for diffusion models have emerged. However, rendering realistic hands remains a challenge, due to the high spatial complexity of hands.  
As shown in Fig. \ref{fig:FIG1}, although Stable Diffusion v1.5~\cite{rombach2022high} with ControlNet~\cite{zhang2023adding} can generate accurate body postures, there are serious distortions in generating hand gestures, such as missing or extra fingers, or non-physiological hand poses. There are two main reasons for generative models failing to generate realistic hand gestures: first, there is a lack of high quality gesture data that can be used by the generative model to learn the spatial structure of the gesture as well as its appearance and the alignment of the hand region with the background environment. Second, there is a lack of methods for learning the complex spatial structures informed by hand priors to guide the generative model.

\begin{figure*}[t]
    \centering
    \includegraphics[width=0.9\textwidth]{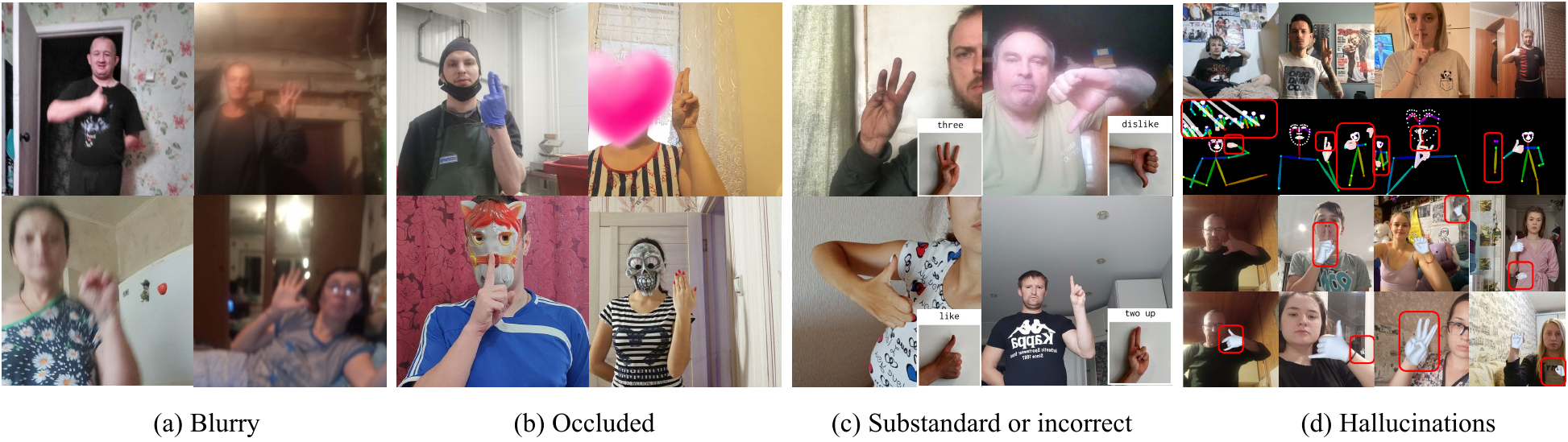}
    \caption{\textbf{Low quality samples in HaGRID~\protect\cite{kapitanov2024hagrid} dataset.} The HaGRID dataset was constructed for gesture recognition tasks only with a hand bounding box and a gesture label, and lacks the necessary annotations for generation tasks. Additionally, the dataset contains many samples with issues such as blurriness, occluded face or hands, substandard gestures or incorrect annotations (compared with standardised gestures in the lower right corner), as well as images that could cause automated annotator hallucinations. Therefore, it is necessary to first clean the HaGRID dataset and then re-annotate it with the labels required for generation tasks.}
    \label{fig:FIG2}
\end{figure*}

The spatial complexity of a hand gesture is reflected in both the endogenous spatial structure of the gesture and its localized environment in the image. We define the data containing the above two types of spatial information as \emph{local context} data. Different from gesture data collected using a greenscreen or related methods \cite{moon2020interhand2}, \cite{moon2024dataset}, local context data refers to the hand image with a rich background. However, there is a lack of suitable datasets: first, relevant datasets either focus on whole-body postures with a small region for hand gestures~\cite{ma2024follow}, or only hand gestures without rich background \cite{moon2020interhand2}, \cite{moon2024dataset}. Such data makes it difficult for the model to learn to synthesize realistic hands that fit with the background. Second, existing datasets lack high-quality spatial annotation, especially 3D structural information. Because of the high  spatial complexity of the hand (especially the fingers), 3D structural annotation is needed to distinguish gestures with similar 2D projections, such as the confusion between palm and back of ``thumbs-up'' gestures shown in Fig. \ref{fig:FIG1}. However, ready-to-use, automatic 3D annotation methods are limited. As shown in Fig. \ref{fig:FIG2}d, even the latest 3D MANO ~\cite{romero2022embodied} mesh annotator HaMeR ~\cite{pavlakos2024reconstructing} can hallucinate, erroneously detecting hands and fingers due to clutter or ambiguities. Therefore, construction of realistic hand synthesis with rich backgrounds remains an open challenge. 

In addition to the scarcity of high-quality local context data, existing hand gesture synthesis improvement methods fall into two categories: \emph{one-stage inference} and \emph{two-stage with inpainting}. The one-stage inference approach uses additional models to supplement hand features~\cite{narasimhaswamy2024handiffuser} or a special loss function~\cite{fu2024adaptive} to train the model to focus on the hand region.  One-stage inference is straightforward, efficient, but challenging because such methods must guide the model to notice a small hand region while learning the whole image, as the hand region has high space complexity.
In contrast to one-stage inference, two-stage inpainting methods divide the complex task into two steps, targeting the malformed hands in generated images. A dedicated model is used for inpainting the hand.  The two-stage with inpainting approach does not require the model to learn hand localization, however these methods face challenges in incorporating the inpainted hand into the whole image.

Regardless of one-stage inference or two-stage with inpainting, existing methods lack the utilization of endogenous spatial structure and environmental localization that guide the generative model to learn gestures along with whole image, resulting in their inability to balance gesture accuracy and realism. To systematically improve the diffusion model's learning of the complex hand spatial structure, we first perform cleansing and re-labeling of the HaGRID datsaet~\cite{kapitanov2024hagrid}, which is the most widely used dataset for gesture-related task, and obtain a high-quality multimodal local context dataset, including the textual caption prompt of the whole image, the depth map of the whole image, the depth map of the hand region, gesture labels, and MANO mesh vertices and the bounding box of the hand region. We re-labeled the dataset using the latest annotator with filtering and alignment. 

We then propose HanDrawer, which focuses on extracting localized and structural information of hand gestures, reconstructing hand regions in a one-stage gesture inference pipeline. Specifically, during training, HanDrawer uses three annotations that characterize the hand region from different perspectives. The first annotation is the gesture label, used for high-dimensional semantic alignment with the text caption prompt used by the diffusion model. The second annotation is the depth of the hand region, which is used to align with the hand region in the whole depth map input used by ControlNet. The third annotation consists of MANO mesh vertices and the hand region bounding box, for learning the endogenous structure information of the gesture and the environment localization of the hand region. HanDrawer learns the alignment and localization embedded within the three modalities mentioned above, as well as the inherent structured information of the gesture. To ensure that the spatially fused features extracted by HanDrawer are used to enhance the corresponding region of the feature map, we perform a position-preserving zero padding (PPZP) fusion strategy.  Specifically, the HanDrawer-extracted features are zero padded according to a hand bounding box in each spatial scale. It then integrates these zero padded features into specific layers of the UNet in the diffusion model, guiding the UNet to pay attention to the hand region during the denoising process to generate more realistic and accurate gestures.

Our contributions are summarized as follows:
\begin{itemize}
    \item We provide an enhanced dataset for hand gesture generation tasks by cleansing and re-annotating the HaGRID v1~\cite{kapitanov2024hagrid} dataset using the latest automated annotators. After manual selection and alignment check, we obtained 28,814 pairs of high-quality multimodal training samples and 7,623 testing samples, covering all the 18 types of gestures in the dataset.

    \item We propose HanDrawer, a unified pipeline suitable for one-stage gesture inference. Specifically, HanDrawer consists of hand region localization and structural feature extraction to guide the diffusion model during the denoising process, enabling the direct generation of realistic and accurate gestures in one-stage inference.  
    
    \item Our proposed position-preserving zero padding (PPZP) fusion strategy ensures feature fusion in the region of interest, enhancing the hand region while maintaining overall image quality.
    
    \item Our experimental validation demonstrates that our proposed method achieves state-of-the-art results for gesture generation quality in quantitative analyses.

\end{itemize}

\section{Related Work}

\subsection{Controllable Human-Centric Image Generation}
Image generation of humans in various poses is a popular task. Recent literature has focused on controlling the human body pose. ControlNet~\cite{zhang2023adding} and T2I-Adapter~\cite{mou2024t2i} enable conditional human-centric image generation by incorporating control signals such as skeletons, sketches, and depth maps.
The Person Image Diffusion Model (PIDM)~\cite{bhunia2023person} integrates pose information and appearance style features through a texture diffusion module during training and utilizes disentangled classifier-free guidance during inference to decouple pose and appearance. HumanSD~\cite{ju2023humansd} demonstrates both experimentally and theoretically that concatenating features of control information (e.g., skeletons) as inputs to a diffusion model and directly fine-tuning the model with sufficient data can yield better results. It also introduced a heatmap-guided denoising loss to enhance the model's focus on humans during inference, effectively mitigating catastrophic forgetting. 
However, the aforementioned human-centric image generation methods overlook the realism and accuracy of hand gesture generation, resulting in distorted or unrealistic hands.

\subsection{Photo-Realistic Hand Gesture Generation}
Existing methods focusing on improved hand generation quality can be broadly categorized into two types: One-stage methods~\cite{narasimhaswamy2024handiffuser},~\cite{fu2024adaptive} and Multi-stage methods~\cite{lu2024handrefiner},~\cite{pelykh2024giving},~\cite{wang2024realishuman},~\cite{zhu2024mole}.

\noindent \textbf{One-stage Methods:} The methods generate an entire image while guiding the model to learn fine-grained features of the hand region, thereby improving the quality of hand region synthesis. Specifically, HandDiffuser~\cite{narasimhaswamy2024handiffuser} enhances the diffusion model's understanding of the intrinsic structure of hand gestures by incorporating 3D mesh information through a dedicated encoder. However, it lacks contextual information for hand localization within the background and requires extensive high-quality annotated data for training. On the other hand, RACL~\cite{fu2024adaptive} employs a specialized loss to train the diffusion model, improving hand pose accuracy while maintaining the overall correctness of the body pose. Nevertheless, RACL shows limited effectiveness in improving the appearance of hand regions. In summary, generating high-quality hand gestures in one-stage inference remains challenging, as it requires the model to balance overall image quality with the complex spatial details of the hand. 

\noindent \textbf{Multi-stage Methods:} These methods decompose the complex generation task into multiple steps, where each step employs a dedicated generative model to synthesize specific regions, and the outputs of these stages are subsequently integrated. The core operation of multi-stage inference is localized inpainting: based on the image generated in the first stage, a customized inpainting model refines the hand region. Compared to one-stage inference, multi-stage inference improves the accuracy and realism of hand generation by increasing inference complexity and the number of models involved.
HandRefiner~\cite{lu2024handrefiner} fine-tunes ControlNet~\cite{zhang2023adding} with hand depth maps to enable the diffusion inpainting model to focus on the hand region. However, the consistency between the hand region inpainted in the second stage and the image generated in the first stage remains suboptimal. 
Recently, RealisHuman~\cite{wang2024realishuman} introduced a multi-condition-controlled diffusion model to generate realistic faces and hands, followed by a second-stage process to repaint areas between the background and the rectified human parts. RealisHuman employs the segmentation model DINO v2~\cite{oquab2023dinov2} to separate faces and hands from the background, enabling precise refinements in these regions of interest. However, due to the lack of background information, RealisHuman does not enable the model to learn the spatial context of these regions.
More recently, MoLE ~\cite{zhu2024mole} trains specialized expert networks for different regions of interest, such as hands and faces, and leverages Soft Mixture Assignment to improve generation quality for these areas. This work also introduces a human-centric dataset for this task. However, MoLE requires additional training stages for each expert network, significantly increasing the training cost.
Unlike the aforementioned one-stage and multi-stage approaches, our proposed HanDrawer equips the diffusion model with the ability to learn complex spatial features by extracting both the local context of the hand region and the intrinsic structural features of hand gestures. As a result, it achieves high-quality hand generation during the denoising process in a single-stage training and inference setup, eliminating the need for multi-stage processes.

\section{HanDrawer}
\label{sec:HanDrawer}

\begin{figure}[t]
    \centering
    \includegraphics[width=\columnwidth]{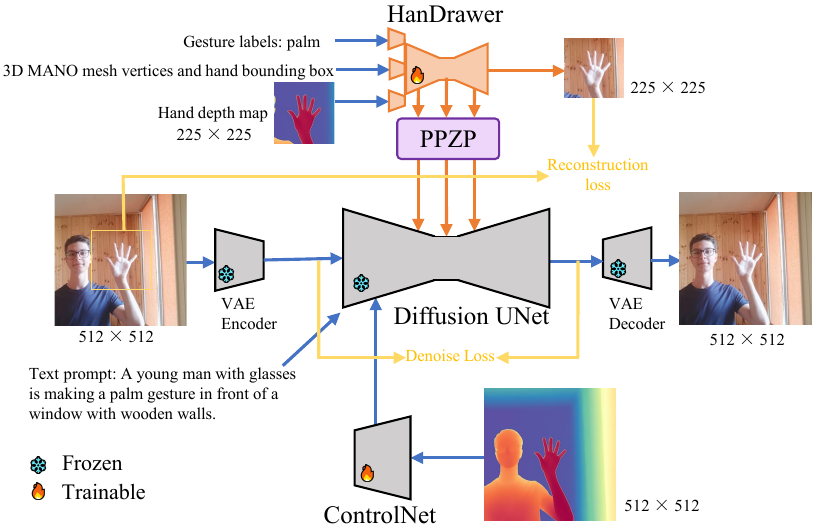}
    \caption{\textbf{HanDrawer training pipeline.} HanDrawer and ControlNet are jointly trained using a reconstruction loss and a denoising loss.}
    \label{fig:FIG3-a}
\end{figure}

\begin{figure}[t]
    \centering
    \includegraphics[width=\columnwidth]{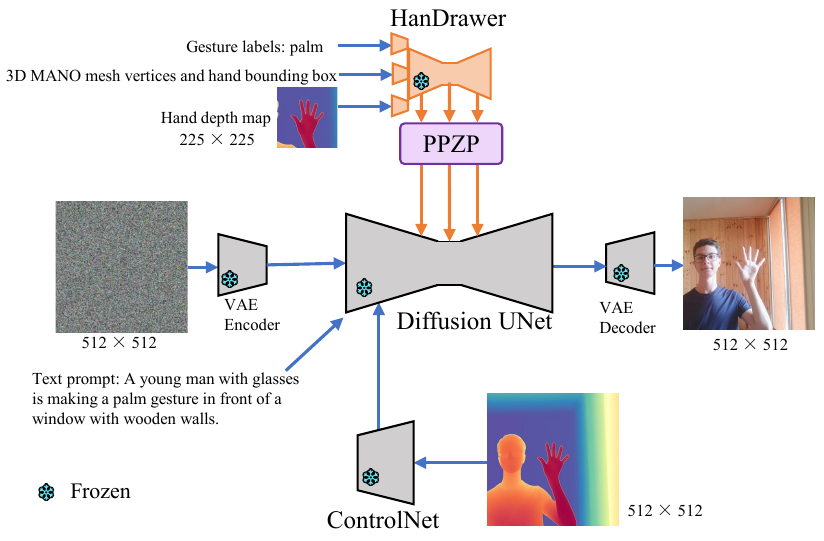}
    \caption{\textbf{HanDrawer inference pipeline.} HanDrawer extracts intrinsic structural features of gestures and region localization from gesture labels, hand region depth maps, and MANO mesh parameters. These spatial features are then fused with appearance features and fed into specific layers of the diffusion model.}
    \label{fig:FIG3-b}
\end{figure}

Fig. \ref{fig:FIG3-a} shows the proposed model training pipeline, where the parameters of the diffusion model and VAE are fixed, while HanDrawer and ControlNet are trained simultaneously. Specifically, the text prompt features, which capture high-level semantic information of the image, and the layout features of the entire image extracted from the depth map by ControlNet are fed into the diffusion model to control the global denoising process. Meanwhile, the corresponding hand gesture labels, depth map, 3D MANO mesh vertices, and 2D hand bounding box are fed into the proposed HanDrawer. HanDrawer extracts the intrinsic spatial structure of the gesture and the contextual localization information of the hand region from multimodal features, fusing these spatial features with the featuremaps of specific layers in the diffusion model, which are described in detail in subsection 3.3. This process guides the diffusion model to focus on the hand region during global denoising, enabling the generation of realistic and accurate hand gestures. Fig. \ref{fig:FIG3-b} shows the proposed inference pipeline.



\subsection{Preliminaries}

\subsubsection{Latent Diffusion Model}
We employ a latent diffusion model with powerful generative capabilities as the foundational generative model.
Latent Diffusion Models (LDMs) ~\cite{rombach2022high} operate in the latent space to reduce the computational cost of high-dimensional data generation. The forward diffusion process gradually adds Gaussian noise to a latent variable $\mathbf{z}_0$, resulting in $\mathbf{z}_t$ defined as:
\[
\mathbf{z}_t = \sqrt{\bar{\alpha}_t} \mathbf{z}_0 + \sqrt{1 - \bar{\alpha}_t} \boldsymbol{\epsilon}, \quad \boldsymbol{\epsilon} \sim \mathcal{N}(\mathbf{0}, \mathbf{I}),
\]
where $\bar{\alpha}_t$ controls the noise schedule. The reverse diffusion process learns to denoise $\mathbf{z}_t$ back to $\mathbf{z}_0$ using a neural network $\epsilon_\theta$. The training objective minimizes the mean squared error (MSE) between the added noise $\boldsymbol{\epsilon}$ and the model's predicted noise:
\[
\mathcal{L}_{\text{LDM}} = \mathbb{E}_{\mathbf{x}, \boldsymbol{\epsilon}, t} \left[ \|\boldsymbol{\epsilon} - \epsilon_\theta(\mathbf{z}_t, t)\|_2^2 \right],
\]
where $\mathbf{z}_t$ is derived from the encoder output of the input data $\mathbf{x}$. This noise prediction task ensures accurate sampling in the reverse diffusion, producing high-quality latent representations that are decoded back into the original data space. LDMs achieve computational efficiency while maintaining the generative performance of diffusion models.

\subsubsection{ControlNet}
Since ControlNet ~\cite{zhang2023adding} has layout control capability, we adopt it as the control module for the whole image.
ControlNet is an extension of diffusion models, designed to enable precise control over the generated content by conditioning on additional structural inputs. It enhances a pretrained diffusion model by introducing a parallel trainable branch that processes control signals, such as skeleton, depth maps, or semantic masks, alongside the original input. The model learns to guide the denoising process by incorporating control features at each diffusion step while retaining the generative quality of the base model. The ControlNet loss is a conditional version of the latent diffusion loss:
\begin{equation}
  {{\cal L}_{{\rm{denoise}}} } = {\mathbb{E}_{\mathbf{x}, \boldsymbol{\epsilon}, t, c} }\left[ {\left\| {\boldsymbol{\epsilon}  - \epsilon_\theta(\mathbf{z}_t, t, c) } \right\|_2^2} \right],
  \label{eq:2}
\end{equation}
where $c$ represents the embeddings of control signals.

\subsubsection{MANO hand model}
We employ MANO  ~\cite{romero2022embodied}, a SMPL-style  ~\cite{loper2023smpl} hand model, to represent the 3D information of hand gestures. The MANO model generates a 3D hand mesh comprising 778 vertices, where each vertex is defined by 3D coordinates \((x, y, z)\). Additionally, the MANO model specifies 21 keypoints corresponding to the anatomical joint positions of the hand, including the wrist and four joints for each finger. The coordinates of these keypoints, \(\mathbf{p} \in \mathbb{R}^{21 \times 3}\), can be computed from the mesh vertices, \(\mathbf{v} \in \mathbb{R}^{778 \times 3}\), through a predefined mapping:
\[
\mathbf{p}_i = \sum_{j=1}^{778} w_{ij} \mathbf{v}_j,
\]
where \(w_{ij}\) represents the mapping weights.
We processed the HaGRID~\cite{kapitanov2024hagrid} dataset using the latest hand MANO mesh recovery tool named HaMeR~\cite{pavlakos2024reconstructing} to obtain the MANO parameters for each gesture as well as the bounding box encompassing the hand.

\subsection{Dataset Curation}

\begin{table}[t!]
\centering
\resizebox{\columnwidth}{!}{
\begin{tabular}{|l|cc|cccc|}
\hline
\multicolumn{1}{|c|}{} & \multicolumn{2}{c|}{Whole Image} & \multicolumn{4}{c|}{Hand Region} \\ \hline
\hline
Dataset  & Image Caption & Depth & Bounding Box & Gesture Label & Depth & MANO Mesh \\
\hline\hline
    HaGRID ~\cite{kapitanov2024hagrid} & $\times$ & $\times$ & \checkmark & \checkmark & $\times$ & $\times$\\
    Local Context Data (Ours) & \checkmark & \checkmark & \checkmark & \checkmark & \checkmark & \checkmark\\
\hline
\end{tabular}}
\caption{\textbf{Annotation comparison of our constructed local context dataset with original HaGRID dataset}. The local context dataset we constructed is not only richer in terms of labeling, which accurately expresses the complex spatial information of the hand, but importantly the whole image annotations and hand region annotations are spatially aligned in the corresponding modalities, facilitating model learning gestures that have a spatial information relationship with the original whole image.}\label{tab:table_dataset}
\end{table}

\begin{figure*}[t]
    \centering
    \includegraphics[width=0.9\textwidth]{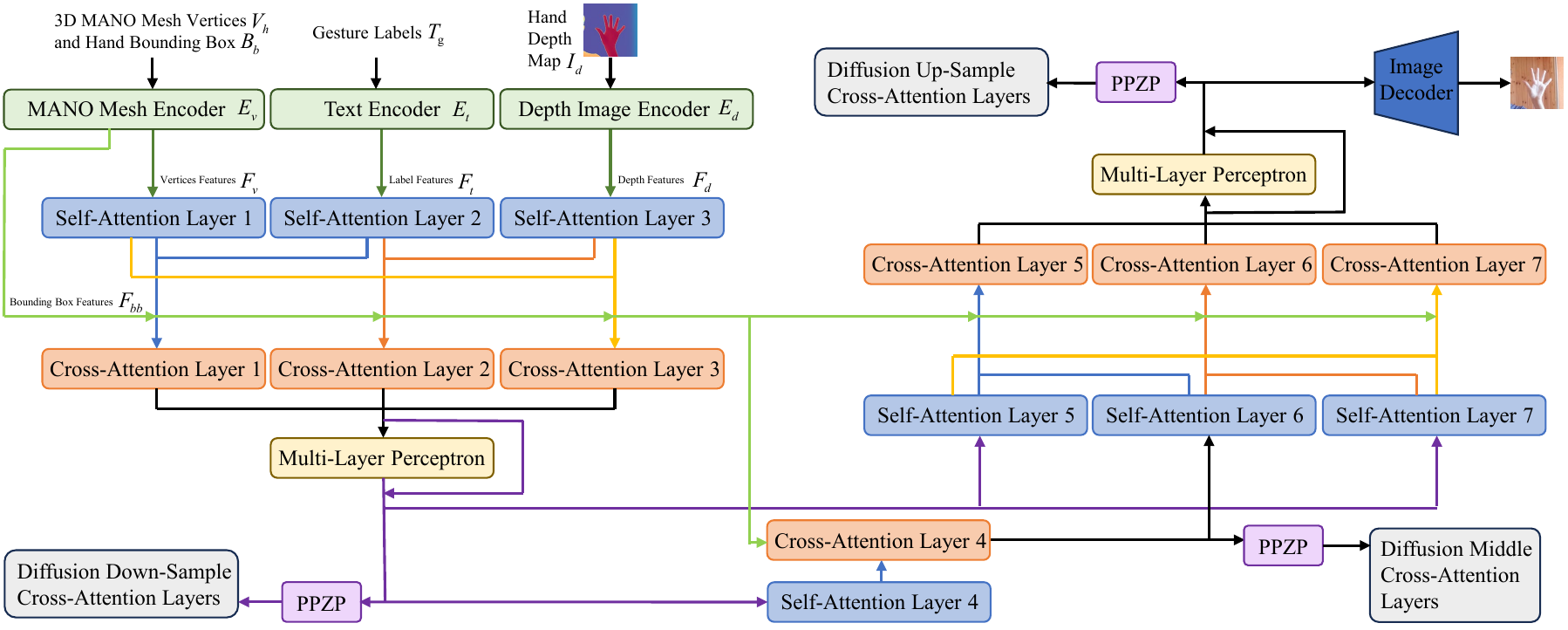}
    \caption{\textbf{HanDrawer architecture.} HanDrawer extracts features of three modalities through self-attention layers and fuses spatial features with these features via cross-attention layers. The gray cross-attention layers belong to the diffusion model. The spatially fused multimodal features extracted by HanDrawer enable the diffusion model to gain spatial understanding of the hand region. While extracting features, HanDrawer reconstructs the hand region, introducing a reconstruction loss to direct the model’s focus to the hand region.}
    \label{fig:FIG4}
\end{figure*}

\begin{figure}[t]
    \centering
    \includegraphics[width=0.8\columnwidth]{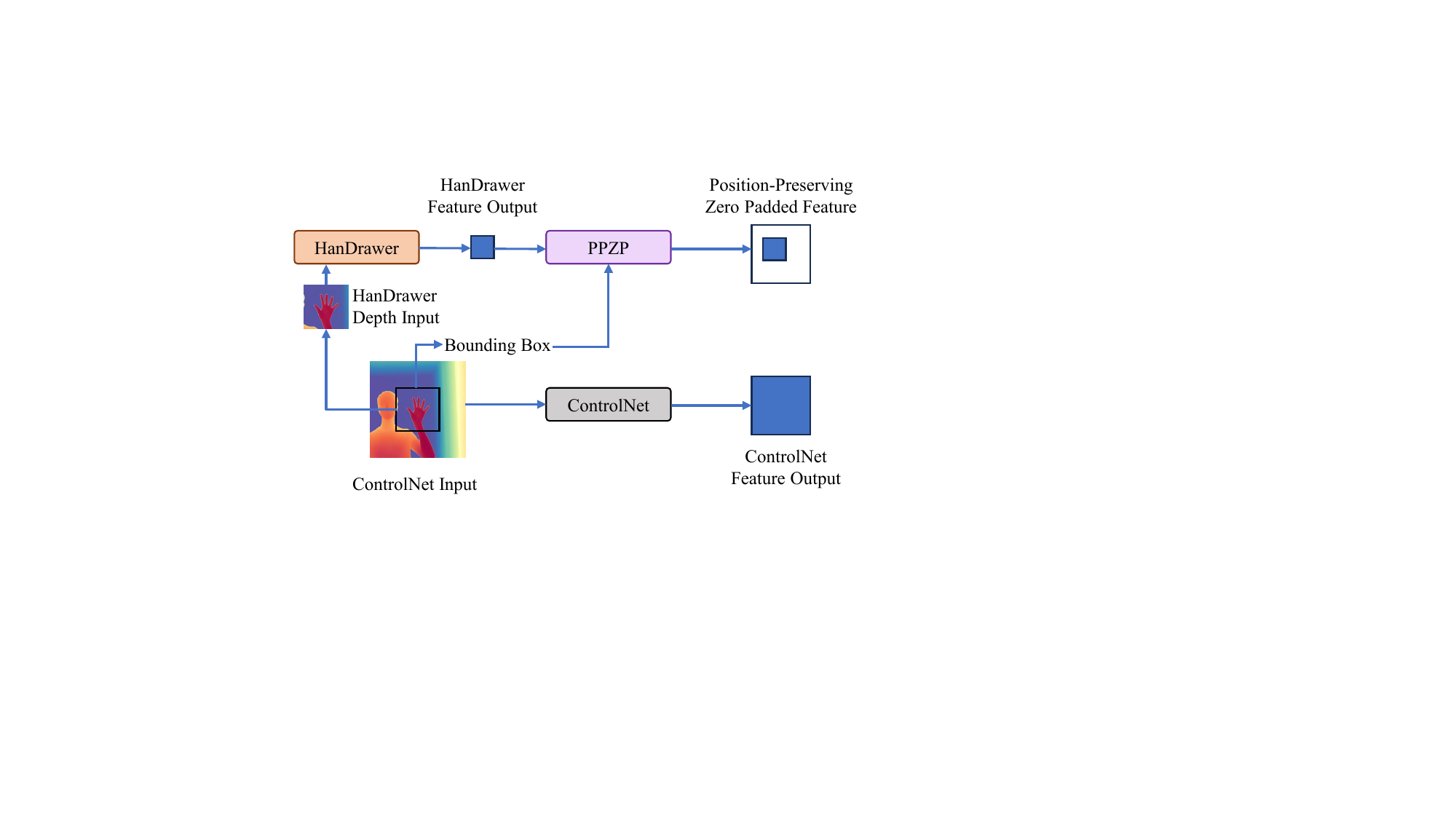}
    \caption{\textbf{Position-preserving zero padding fusion strategy.} By applying position-preserving zero padding to the hand region features extracted by HanDrawer based on the normalized hand region bounding box, we ensure the extracted hand features are only used to refine the hand region features in the UNet feature map without affecting the features of other regions.}
    \label{fig:FIG5}
\end{figure}

The HaGRID dataset is currently a widely used dataset for gesture-related tasks. However, since the original dataset was designed for gesture recognition tasks, it lacks text caption prompts and conditional control required for posture-controlled generation tasks. Additionally, the dataset consists of user-collected, in-the-wild data, resulting in quality fluctuations and significant noise. Specifically, as shown in Fig. \ref{fig:FIG2}, the HaGRID dataset has problematic samples, including images with exposure issues, motion blur causing image distortion, occluded facial or hands, non-standard or mislabeled gestures.  There are also samples that produce annotation errors (e.g., hallucinations) due to clutter or ambiguity. Such low-quality or incorrect samples are detrimental to generation tasks, making it essential to clean the dataset thoroughly and select high-quality image data for re-annotation.

Inspired by recent work~\cite{fu2024adaptive}, we cropped $512\times512$ images centered on the hands and used the annotation tool DWPose~\cite{yang2023effective} to preliminarily filter these cropped images, removing samples prone to errors like multiple heads, multiple hands, or distortions. Next, we used the latest depth annotation tool DepthAnyVideo~\cite{yang2024depth} and the 3D MANO mesh annotation tool HaMeR to annotate the filtered data, obtaining depth maps, mesh vertices, and bounding boxes for the hand regions. Based on these bounding boxes, we then cropped corresponding $225\times225$  original and depth images, centered on the hand region. This $225\times225$ crop size focuses on gestures while retaining some contextual information, which is essential for the model to learn the localization information of the hand regions.

Additionally, to obtain text caption prompts for each image, we designed a prompt engineering approach that guides LLaVA v1.5~\cite{liu2024improved} to annotate each image with gesture and background text based on its original gesture label. This ensures feature alignment between the proposed HanDrawer and the diffusion model at the semantic level. Finally, we performed manual filtering and alignment on the resulting multimodal data, yielding 28,814 pairs of high-quality training samples and 7,623 pairs of high-quality test samples with annotations across three modalities: text, depth, and 3D MANO mesh vertices with hand bounding boxes, covering all 18 gestures from HaGRID v1. Table \ref{tab:table_dataset} shows the annotation comparison of our constructed local context dataset with original HaGRID dataset. 

\subsection{HanDrawer}

Fig.~\ref{fig:FIG4} illustrates the structure of the proposed HanDrawer. Overall, this is a multimodal generative UNet model built with attention layers, with three input modalities: gesture labels $T_g$, hand region depth $I_{d}$, and MANO mesh vertices $V_h$ with 2D hand bounding box $B_b$. The model has two output modalities: the reconstructed image of the hand region and multilayer spatially fused feature maps.

Specifically, in the encoder $E$, three separate tiny encoders, namely $E_t$, $E_{d}$, and $E_v$, 
extract the text features $F_t$,  the depth features $F_{d}$, the vertex features $F_{v}$, and the bounding box features $F_{bb}$ from the three input modalities, 
each fed into its respective Self-Attention Layer~\cite{vaswani2017attention}.  The text encoder $E_t$ is same as the diffusion text encoder. The depth encoder $E_{d}$ employs two convolutional embedding modules tailored for single-channel and multi-channel depth images, producing low-dimensional feature representations with positional embeddings. And the graph-based mesh vertices encoder $E_v$ combines a three-layer graph convolutional network to extract 3D mesh vertex features and a multi-head attention mechanism to process bounding box information, resulting in normalized global embeddings. This design effectively captures spatial and geometric features from the input image and mesh vertices.
Notably, the feature vectors $F_{bb}$ from the Hand Bounding Box are fed into all Cross-Attention Layers in HanDrawer to provide hand localization information during the multimodal alignment and fusion process. The outputs of the Self-Attention Layers (labelled 1, 2, 3 in Fig.~\ref{fig:FIG4}) in the down-sampling block are paired and, together with the Bounding Box Features $F_{bb}$, and input to the three Cross-Attention Layers in the down-sampling area for feature alignment and fusion. The outputs from the Cross-Attention Layers are then merged via a Multi-Layer Perceptron (MLP) module and passed to the Self-Attention Layers of HanDrawer’s middle block and up-sample block, as well as to the Cross-Attention Layers in the diffusion model’s down-sample block.

In HanDrawer’s middle block, the output from the Self-Attention Layer (labelled 4 in Fig.~\ref{fig:FIG4}), along with the Bounding Box Features $F_{bb}$, are input to the Cross-Attention Layer  to produce the middle block’s feature output. This middle block feature output is passed to the Cross-Attention Layers in the diffusion model’s middle block and to a Self-Attention Layer in the HanDrawer’s up-sample block. The other two Self-Attention Layers in HanDrawer’s up-sample block receive skip connection outputs from the down-sample block. The outputs of these Self-Attention Layers (labelled 5, 6, and 7 in Fig.~\ref{fig:FIG4}) are paired, combined with Bounding Box Features $F_{bb}$, and then passed to the three Cross-Attention Layers in the up-sample block for feature alignment and fusion. Finally, the outputs of the three Cross-Attention Layers are merged through an MLP to generate the up-sample block’s feature output. This feature output is fed into the Cross-Attention Layers in the diffusion model’s up-sample block and into HanDrawer’s image decoder. HanDrawer’s image decoder reconstructs the hand region, while all fused features fed into the diffusion model guide the diffusion model to pay attention to the hand region and generate realistic and accurate hand gestures during the denoising process.

As shown in Fig.~\ref{fig:FIG5}, to ensure the spatial correspondence between the hand region features $F_{HanDrawer}$ extracted by HanDrawer and the intermediate layer features $F_{UNet}$ of UNet, we first apply PPZP to the hand region feature map based on the normalized hand region bounding box before feature fusion. In these zero padded feature maps of HanDrawer, only the region corresponding to the hand region in the original image (scaled proportionally) contains non-zero feature values, while all other regions are set to zero. These zero padded feature maps of HanDrawer are then added to the corresponding intermediate layer feature map of UNet to obtain the fused features.

\subsection{Training}

The denoising loss used in latent diffusion models focuses on the entire image. To guide the model in learning local features of the hand region, we introduce a reconstruction loss targeted for the hand region to train the entire pipeline. Specifically, we calculate the reconstruction loss between the hand region ${\hat {\bf I}}_{\rm{reHand}}$ generated by HanDrawer and the corresponding hand region in the original image ${\bf I}_{{\rm{reHand}}}$ as follows:
\begin{equation} \label{E4}
\begin{aligned}
{{{\cal L}_{{\rm{reHand}}}} =} &{\mathbb E} \left| {{\bf I}_{{\rm{reHand}}} - \hat {\bf I}}_{{\rm{reHand}}} \right|,
\end{aligned}
\end{equation}
Then, the reconstruction loss and denoising loss are added to obtain the overall loss as follows:
\begin{equation} \label{E4}
\begin{aligned}
{\cal L} = {\cal L}_{{\rm{denoise}}} + \lambda {\cal L}_{{\rm{reHand}}},
\end{aligned}
\end{equation}
where $\lambda$ is a weight to balance ${{\cal L}_{\rm{denoise}}}$ and ${{\cal L}_{\rm{reHand}}}$.

\section{Experiments}
\label{sec:experiments}


\begin{table}[t!]
\centering
\resizebox{\columnwidth}{!}{
\begin{tabular}{|l|cccc|}
\hline
Method  & FID ↓ & KID ↓ & FID-Hand ↓ & KID-Hand ↓ \\
\hline\hline
    SD v1.5 + ControlNet~\cite{zhang2023adding} & 31.4976 & 0.0238$\pm$0.0002 & 32.2157 & 0.0238$\pm$0.0007\\
    HandRefiner~\cite{lu2024handrefiner} & 35.7291 & 0.0296$\pm$0.0008 & 35.4393 & 0.0259$\pm$0.0004\\
    RealisHuman~\cite{wang2024realishuman} & 31.0369 & 0.0232$\pm$0.0001 & 30.2902 & 0.0210$\pm$0.0004\\
    HanDrawer (Ours) & \textbf{26.8279} & \textbf{0.0196$\pm$0.0002} & \textbf{28.7506} & \textbf{0.0201$\pm$0.0008}\\
\hline
\end{tabular}}
\caption{Quantitative results.  It can be seen that our proposed HanDrawer has a substantial improvement in all metrics compared to competing models. Best quality is highlighted in \textbf{bold}.}\label{tab:table1}
\end{table}

\begin{figure*}[t]
    \centering
    \includegraphics[width=\textwidth]{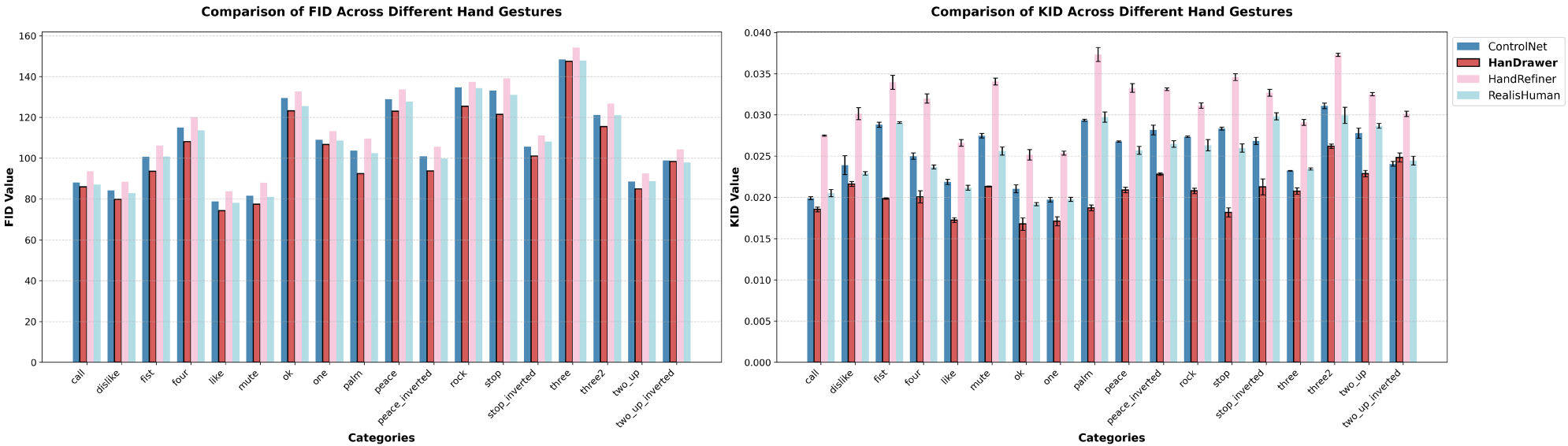}
    \caption{\textbf{The performance comparison of different methods on different gestures.} Our proposed HanDrawer outperforms the baseline methods in terms of FID and KID.}
    \label{fig:FIG8}
\end{figure*}

\begin{table}[t!]
\centering
\resizebox{\columnwidth}{!}{
\begin{tabular}{|l|cccc|}
\hline
Method  & FID ↓ & KID ↓ & FID-Hand ↓ & KID-Hand ↓ \\
\hline\hline
    without PPZP without graph & 38.6939 & 0.0317$\pm$0.0004 & 59.6560 & 0.0525$\pm$0.0002 \\
    without graph & 30.7333 & 0.0231$\pm$0.0004 & 52.2445 & 0.0601$\pm$0.0009 \\
    without 3Dmesh & 29.9273 & 0.0224$\pm$0.0004 & 50.9478 & 0.0431$\pm$0.0003 \\
    HanDrawer & \textbf{26.8279} & \textbf{0.0196$\pm$0.0002} & \textbf{28.7506} & \textbf{0.0201$\pm$0.0008} \\
\hline
\end{tabular}}
\caption{Ablation results. It can be seen that the proposed settings are the best solution for comprehensive quality improvement in the hand region. Best quality is highlighted in \textbf{bold}.}\label{tab:table2}
\end{table}

\subsection{Implementation Details}
We first cleaned and relabeled the HaGRID v1 dataset, resulting in 28,814 training pairs and 7,623 test pairs, each annotated with text, depth, and 3D MANO mesh vertices with bounding boxes. These data pairs cover all 18 hand gestures in HaGRID v1. The entire image size is $512 \times 512$, while the gesture region is cropped to $225 \times 225$. We used this data to train and test our proposed model. Specifically, we jointly trained our proposed HanDrawer model with the frozen Stable Diffusion v1.5 model, and the trainable ControlNet model for 90,000 steps, with a learning rate of $1 {e^{ - 6}}$, $\lambda = 0.1$ based on empirical experiments and a batch size of 6. We chose Stable Diffusion v1.5 with ControlNet.  We compare to HandRefiner and RealisHuman, the latest open source multi-stage inference methods. In our experiments, HandRefiner and RealisHuman repair the generated results produced by Stable Diffusion v1.5 with ControlNet. All training and inference were performed on a single NVIDIA Tesla 80GB-A100 GPU.

\subsection{Evaluation Metrics}

To maintain consistency with previous work, we used Frechet Inception Distance (FID)~\cite{heusel2017gans} and Kernel Inception Distance (KID)~\cite{binkowski2018demystifying} to evaluate the generated full images, and FID-H and KID-H to assess the hand region in the generated images. Since the Inception v3 model for calculating FID requires a  $299\times299$  size image, we crop the image with $299\times299$ size centred on the hand bounding box to get the image of the hand region.


\subsection{Quantitative Results}
The quantitative results are presented in Table \ref{tab:table1}. Our proposed HanDrawer method outperforms competing methods in all metrics by a substantial margin. This is because our method includes additional hand-related modalities, extracts complex spatial information from the hand region and uses these features to modify the hand region in the feature map of the diffusion model. More importantly, our proposed PPZP leverages the hand region localization information to ensure that the process from feature extraction to feature fusion is position-preserving, so that the hand region is enhanced without affecting other regions.

Fig.~\ref{fig:FIG8} compares performance for different gestures. It can be seen that our proposed HanDrawer outperforms the competing methods in terms of FID and KID on most of the gestures except for the gesture "two up inverted".

\subsection{Qualitative Results}
Fig. \ref{fig:FIG1} shows some qualitative results. More qualitative results are placed in the supplementary material.


\subsection{Ablation Studies}

The quantitative ablation results are presented in Table \ref{tab:table2}. Our proposed setting is the best solution for comprehensive quality improvement in the hand area. First, we ablate feature padding and graph based mesh encoder, which means that instead of using zero padding in the feature fusion, the output size of the HanDrawer is set to be the same as the feature size of the corresponding layer of the UNet. By comparing the use and non-use of feature padding, it can be seen that the use of zero padding has a greater performance improvement on all metrics. This is because the use of zero padding ensures that the hand features extracted by HanDrawer are fused into the hand region of the corresponding layer features of UNet. Next, without 3D mesh means that the 3D mesh data is not fed into the HanDrawer. It can be seen that 3D mesh is important for hand-related performance improvement.



\section{Conclusions}
\label{sec:Conclusions}

This paper aims to enhance the realism and accuracy of gesture generation in diffusion models. First, to construct high-quality multimodal local context data, we cleaned and re-annotated the HaGRID v1 dataset. Then, to endow the diffusion model with spatial feature learning capabilities for improving the realism and accuracy of complex gesture generation, we propose HanDrawer, a module that can extract the intrinsic spatial structure of gestures and the localization of hand regions. By integrating these complex spatial features with specific layers of the diffusion model, HanDrawer guides the model to focus on the hand region during the denoising process, enabling realistic and accurate gesture generation. Additionally, HanDrawer reconstructs the hand region while extracting features, introducing a reconstruction loss to train the model to learn the hand region. The simulation verifies from multiple evaluation metrics that the proposed method significantly improves the quality of gesture generation and achieves state-of-the-art in both quantitative and qualitative analyses.

\subsubsection{Ethical Statement}

While the primary goal of HanDrawer is to advance generative AI techniques for high-quality image synthesis with accurate hand rendering, we acknowledge the risk of misuse in creating deceptive or harmful content (e.g., deepfakes).  HanDrawer’s improved ability to render hands has significant potential in areas such as education, virtual reality, gaming, and accessibility technologies. For instance, the technology can aid in creating more lifelike avatars for sign language interpretation.  To promote transparency and reproducibility, the code and pre-trained models will be made available with clear guidelines on usage.  We also commit to engaging with the research community and policymakers to discuss safeguards against unethical applications of generative AI.

\bibliographystyle{named}
\bibliography{ijcai25}

\end{document}